%% file: main.tex
\documentclass[letterpaper, 10 pt, conference]{ieeeconf}
\usepackage{xspace}
\IEEEoverridecommandlockouts
\usepackage[font={footnotesize}]{caption}
\usepackage{graphicx}
\usepackage{amsmath}
\usepackage{tikz}
\usepackage{cite}
\usepackage{bbold}
\usepackage{hyperref}
\usepackage{booktabs}

\newcommand{\secref}[1]{Section~\ref{#1}}
\newcommand{\nam}{navigation-and-manipulation\xspace}

\title{\LARGE \bf
Visibility-Aware Mobile Grasping in Dynamic Environments
}

\author{%
Tianrun~Hu$^{1,2}$, Anxing~Xiao$^{1,2}$, David~Hsu$^{1,2}$, and Hanbo~Zhang$^{3}$%
\thanks{$^{1}$School of Computing, National University of Singapore, Singapore.}%
\thanks{$^{2}$Smart Systems Institute, National University of Singapore, Singapore.}%
\thanks{$^{3}$Shanghai Innovation Institute, Shanghai, China.}%
\thanks{\raggedright Emails: {\tt\small \{tianrun.hu, anxingxiao\}@u.nus.edu}, {\tt\small dyhsu@comp.nus.edu.sg}, {\tt\small zhanghanbo@sii.edu.cn}.}%
}

\hypersetup{
  pdftitle={Visibility-Aware Mobile Grasping in Dynamic Environments},
  pdfauthor={Tianrun Hu, Anxing Xiao, David Hsu, Hanbo Zhang}
}

\begin{document}

\IEEEaftertitletext{%
  \vspace{-0.8em}
  \begin{minipage}{\textwidth}
    \centering
    \includegraphics[width=0.97\textwidth]{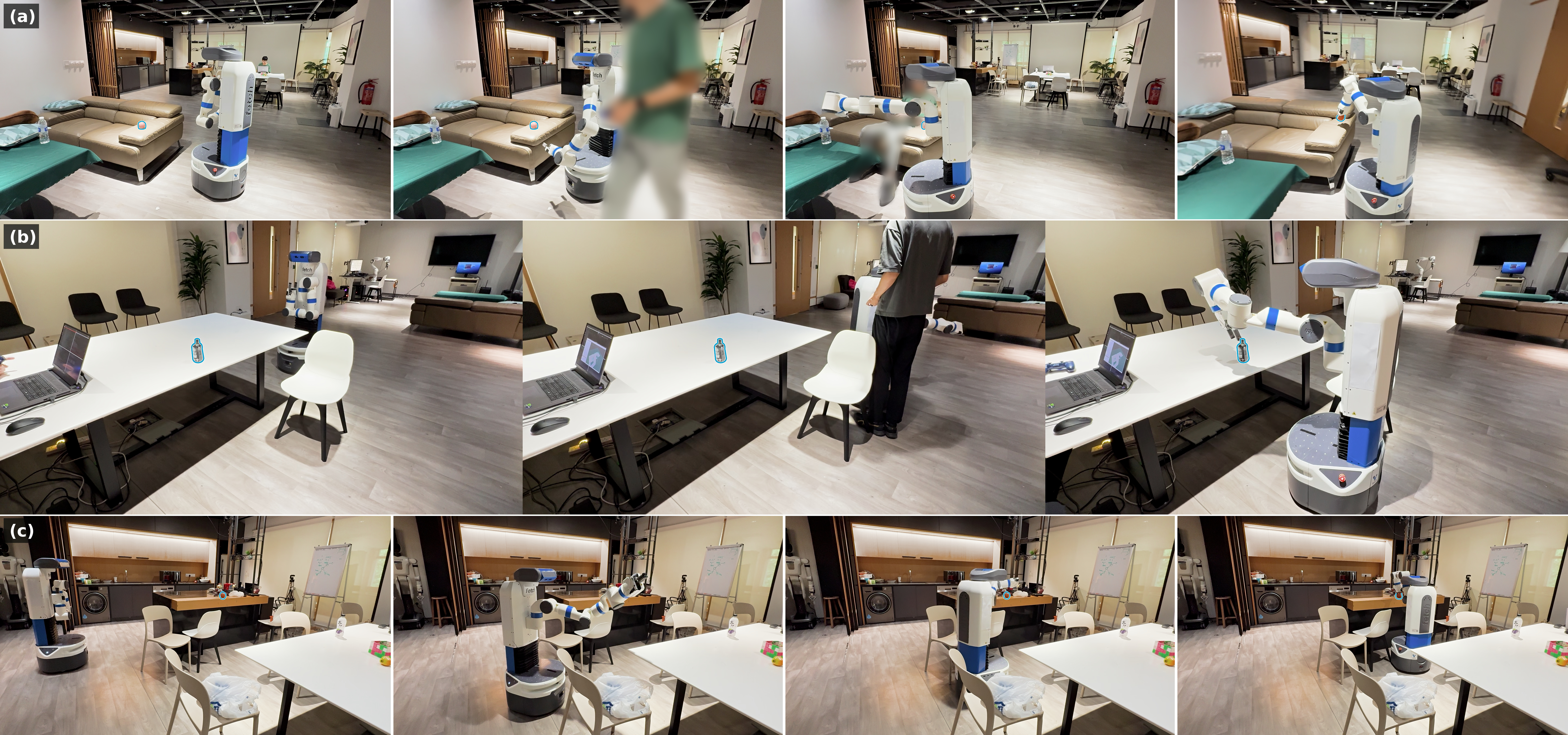}
    \captionof{figure}{\textbf{Visibility-aware mobile grasping in the real world.}
      (a) A moving person enters the robot's route, and the robot reacts before grasping the apple.
      (b) The robot navigates around a person and chair to reach a bottle on a dining table.
      (c) The robot navigates through tightly spaced chairs before reaching an apple on the workstation. Faces are blurred for privacy.}
    \label{fig:problem_setup}
  \end{minipage}
  \vspace{0.5em}
}

\maketitle
\begin{abstract}
We consider mobile grasping in unknown and dynamic environments, where a robot must approach and grasp a target object using a camera with a limited field of view.
This setting introduces two visibility constraints. Incomplete environmental observations can create collision risks along the whole-body trajectory, while occlusions can hinder object pose refinement during grasping.
Previous systems often assume known or static environments and decouple perception from motion, leaving safety unresolved around unobserved obstacles.
We present a unified system that enforces both visibility constraints at runtime through three core components. A velocity-aware gaze policy directs perception toward fast-moving regions of the planned swept volume, a hierarchical subgoal policy samples grasp and intermediate configurations that maintain target visibility, and a real-time whole-body planner replans over a receding horizon as the map updates.
Across 400 randomized simulation scenarios and real-world trials on a Fetch mobile manipulator, our system achieves success rates of 70.4\% in unknown static environments and 57.9\% in dynamic environments, outperforming the decoupled \nam strategy by 16.3 and 9.3 percentage points, respectively, while significantly reducing collisions.
Project page: \url{https://visibility-aware-mobile-grasping.github.io/}.
\end{abstract}

\input{introduction/introduction}
\input{literature_review/literature_review}
\input{formulation/formulation}
\input{method/method}

\input{experiments/experiments}

\input{results/results}

\input{conclusion/conclusion}

\section*{Acknowledgment}
The authors used OpenAI's ChatGPT and Codex to assist with code generation, idea formulation, and manuscript text editing. The authors take full responsibility for the final manuscript, implementation, analyses, and conclusions.

\bibliographystyle{IEEEtran}
\bibliography{bib/references}

\end{document}

%% file: introduction/introduction.tex
\section{Introduction}
\label{sec:introduction}

Bringing mobile robots from structured environments such as factories and warehouses to unstructured environments \cite{liu2024okrobot, yenamandra2024homerobotopenvocabularymobilemanipulation, xiao2025robibutlermultimodalremote} such as homes, offices, and hospitals requires mobile manipulation beyond the common assumptions of a pre-built, perfect map and static obstacles. When the environment contains unknown clutter, such as a backpack left on the floor, or dynamic agents, such as a person walking by (Figure~\ref{fig:problem_setup}), the robot must actively use its onboard sensors to gather critical information while moving. In this paper, we focus on this challenging task regime. We propose a \textit{visibility-aware mobile grasping system for unknown and dynamic environments that explicitly respects visibility constraints throughout the mobile grasping process}.

We formulate the problem as sampling over the constraint graph illustrated in Figure~\ref{fig:visibility_factor_graph}. Mobile grasping involves constraints on collision avoidance, kinematics, and grasping, which existing systems typically address under known maps. A limited field of view in unknown environments introduces two additional visibility constraints. The \textit{collision visibility constraint} requires the maintained map to maximize coverage of the whole-body swept volume. Otherwise, a collision-free trajectory may still hit unobserved obstacles. The \textit{object visibility constraint} requires the grasp configuration to keep the target in view for continued pose refinement. With a single camera, these constraints compete for gaze between the swept volume during motion and the target during grasping. We call this \textit{visibility-aware mobile grasping}. Classical model-based \cite{9830863,6630839} and data-driven systems \cite{yenamandra2024homerobotopenvocabularymobilemanipulation} struggle under such partial observability, while POMDP methods \cite{somani2013despot} are often intractable for high-dimensional whole-body manipulation under real-time requirements.

\input{figures/visibility_factor_graph}

To address the visibility challenges while meeting practical deployment requirements, we present a unified system that enforces both constraints at runtime in a hierarchical and reactive manner. For the \textit{collision visibility constraint}, we propose a velocity-aware active perception policy that continuously directs the camera toward the swept volume of the current trajectory, prioritizing fast-moving and imminent regions. For the \textit{object visibility constraint}, we introduce a hierarchical subgoal policy that selects collision-free whole-body grasp and pre-grasp configurations while keeping the target in view. When manipulation is blocked or fails, the policy reactively falls back to more exploratory subgoals, such as observation poses. Finally, a real-time whole-body motion planner \cite{10611190} and tracker replan within tens of milliseconds as the map updates. Together, these components turn dynamically changing visibility constraints into safe and reactive motion in the high-dimensional configuration space.

To evaluate our system and derive practical design principles for visibility-aware mobile grasping, we conduct an extensive study across 400 randomized simulation scenarios and real-world trials on a Fetch mobile manipulator. The experiments range from static environments with unknown layouts to dynamic scenes where obstacles cross the robot's active route. Our system significantly outperforms the baselines in both unknown static and dynamic environments. The results suggest that interleaving whole-body planning with active perception enables mobile manipulators to operate more effectively in high-dimensional configuration spaces under real-world uncertainty. We further summarize the key design principles in the discussion section and provide a practical improvement direction for long-term open-world deployment of mobile manipulation systems.

%% file: figures/visibility_factor_graph.tex
\usetikzlibrary{shapes}
\begin{figure}[t]
    \centering
    \resizebox{\columnwidth}{!}{%
        \begin{tikzpicture}[
                var/.style={circle, draw=black, thick, fill=white, minimum size=1.05cm, inner sep=1pt, font=\Large},
                subvar/.style={circle, draw=black, thin, fill=white, minimum size=0.82cm, inner sep=0pt, font=\large},
                group/.style={ellipse, draw=black, thick, minimum width=3.1cm, minimum height=1.8cm, inner sep=0pt},
                factor/.style={rectangle, draw=black, thick, minimum size=0.44cm, inner sep=0pt},
                flabel/.style={font=\Large},
                caption/.style={align=center, font=\normalsize, text=black},
                edge/.style={thick}
            ]

            \node[var] (M) at (0,0) {$M_t$};

            \node[group] (qinit) at (2.15,0) {};
            \node[font=\Large] at (2.15,0.56) {$q_t$};
            \node[subvar] at (1.25,-0.26) {$q_{b}$};
            \node[subvar] at (2.15,-0.26) {$q_{ta}$};
            \node[subvar] at (3.05,-0.26) {$q_{v}$};

            \node[var, fill=orange!30] (Twb) at (4.4,0) {$\xi_t$};

            \node[group, fill=blue!12] (qgoal) at (6.65,0) {};
            \node[font=\Large] at (6.65,0.56) {$(g_t,v)$};
            \node[subvar] at (5.75,-0.26) {$q_{b}$};
            \node[subvar] at (6.65,-0.26) {$q_{ta}$};
            \node[subvar] at (7.55,-0.26) {$v$};

            \node[var, fill=green!25] (Tee) at (8.95,0) {$T_{ee}$};
            \node[var] (o) at (10.75,0) {$o$};

            \node[caption, anchor=north] at (0,-0.68) {maintained\\map};
            \node[caption, anchor=north] at (2.15,-1.05) {current\\configuration};
            \node[caption, anchor=north] at (4.4,-0.68) {whole-body\\motion traj};
            \node[caption, anchor=north] at (6.65,-1.05) {goal\\configuration};
            \node[caption, anchor=north] at (8.95,-0.68) {final grasp\\EE pose};
            \node[caption, anchor=north] at (10.75,-0.68) {target\\obj};

            \node[factor, fill=red!80!black] (phicol) at (1.7,3.0) {};
            \node[flabel, text=red!80!black, anchor=south] at (1.7,3.28) {$\phi_{col}$};
            \node[factor, fill=violet] (phimotion) at (3.9,3.0) {};
            \node[flabel, text=violet, anchor=south] at (3.9,3.28) {$\phi_{motion}$};
            \node[factor, fill=green!45!black] (phiik) at (6.4,3.0) {};
            \node[flabel, text=green!45!black, anchor=south] at (6.4,3.28) {$\phi_{IK}$};
            \node[factor, fill=olive] (phigrasp) at (9.3,3.0) {};
            \node[flabel, text=olive, anchor=south] at (9.3,3.28) {$\phi_{grasp}$};

            \node[factor, fill=orange!80!black] (phicolvis) at (2.2,-3.0) {};
            \node[flabel, text=orange!80!black, anchor=west] at (2.5,-3.0) {$\phi_{colvis}$};
            \node[caption, anchor=north] at (2.2,-3.35) {collision visibility\\constraint};
            \node[factor, fill=blue!70!black] (phiobjvis) at (6.6,-3.0) {};
            \node[flabel, text=blue!70!black, anchor=east] at (6.3,-3.0) {$\phi_{objvis}$};
            \node[caption, anchor=north] at (6.6,-3.35) {object visibility\\constraint};

            \node[anchor=north east, align=left, font=\normalsize, draw=black!40, rounded corners=2pt, inner sep=5pt] at (11.3,-2.4) {%
                $q_{b}$ \ base\\
                $q_{ta}$ torso\&arm\\
                $q_v, v$ \ head};

            \node[font=\normalsize, rotate=90, anchor=south] at (-0.9,3.0) {constraints};
            \node[font=\normalsize, rotate=90, anchor=south] at (-0.9,0) {variables};

            \draw[edge, red!80!black] (phicol) to[out=-160, in=80] (M);
            \draw[edge, red!80!black] (phicol) to[out=-110, in=75] (qinit);
            \draw[edge, red!80!black] (phicol) to[out=-30, in=110] (Twb);
            \draw[edge, red!80!black] (phicol) to[out=-8, in=140] (Tee);

            \draw[edge, violet] (phimotion) to[out=-140, in=70] (qinit);
            \draw[edge, violet] (phimotion) to[out=-80, in=95] (Twb);
            \draw[edge, violet] (phimotion) to[out=-30, in=110] (qgoal);

            \draw[edge, green!45!black] (phiik) to[out=-120, in=70] (qgoal);
            \draw[edge, green!45!black] (phiik) to[out=-50, in=110] (Tee);

            \draw[edge, olive] (phigrasp) to[out=-115, in=75] (Tee);
            \draw[edge, olive] (phigrasp) to[out=-55, in=95] (o);

            \draw[edge, orange!80!black] (phicolvis) to[out=170, in=-75] (M);
            \draw[edge, orange!80!black] (phicolvis) to[out=15, in=-105] (Twb);

            \draw[edge, blue!70!black] (phiobjvis) to[out=120, in=-90] (qgoal);
            \draw[edge, blue!70!black] (phiobjvis) to[out=20, in=-110] (o);

        \end{tikzpicture}%
    }
    \caption{\textbf{Constraint graph of visibility-aware mobile grasping.}
        Circles denote variables and squares denote constraints. Beyond collision $\phi_{col}$, motion $\phi_{motion}$, IK $\phi_{IK}$, and grasping $\phi_{grasp}$ constraints, $\phi_{colvis}$ requires refreshed observation coverage of the swept volume of $\xi_t$, while $\phi_{objvis}$ requires goal $g_t$ with head orientation $v$ to keep target object $o$ at $p$ in view.
        }
    \label{fig:visibility_factor_graph}
    \vspace{-0.5cm}
\end{figure}

%% file: literature_review/literature_review.tex
\section{Related Work}
\label{sec:related_work}

\subsection{Mobile manipulation systems.}
Deployed mobile manipulation systems operate reliably in controlled settings such as retail environments and long-term human-shared spaces \cite{Bajracharya_2023, 9517662, xiao2025robibutlermultimodalremote, xiao2026hume, yu2026canine}, but they depend on pre-built 3D maps and detect dynamic obstacles with 2D lidar, so they cannot react safely to changes in rich 3D structure.
Reactive planners that address partial observability and dynamic obstacles \cite{Covic_2025, Finean_2021} are validated mostly in simplified simulations or low-dimensional configuration spaces, leaving real-time whole-body safety in cluttered, dynamic 3D scenes unaddressed.
To manage the high dimensionality of mobile manipulation, prevailing pipelines decouple the problem through base placement optimization, selecting a fixed configuration that maximizes manipulability for a known goal~\cite{9830863, 6630839}.
This assumes a nearly complete environment model and a goal fixed upfront.
Our hierarchical subgoal generation extends this idea to partial observability, refining intermediate configurations incrementally as the map evolves.

\medskip
\subsection{Visibility-aware planning and active perception.}
Visibility-aware planners constrain motion to previously observed free space~\cite{goretkin2018look, ichter2017perceptionawaremotionplanningmultiobjective} under the assumption that observed regions remain valid in a static environment.
Joint optimization of gaze and motion~\cite{meng2025lookleapplanningsimultaneous} maintains visibility of a fixed manipulation goal rather than one refined online.
Active perception takes the complementary strategy of selecting viewpoints to maximize information gain.
Frontier-based exploration builds complete maps~\cite{4209173}, next-best-view planning optimizes viewpoints for grasping~\cite{breyer2022closedloopnextbestviewplanningtargetdriven, 10.1109/ICRA.2019.8793805}, and recent work couples perception with manipulation to reduce occlusion~\cite{jauhri2024activeperceptivemotiongenerationmobile, marques2025mapspacebeliefprediction}.
These methods separate viewpoint selection from motion execution and do not reason about swept-volume safety during replanning.
Unlike them, our gaze policy weights observation by planned velocity over the robot's future swept volume, so perception serves collision monitoring and goal refinement simultaneously.
Real-time replanning itself builds on advances in reactive control~\cite{Haviland_2021}, optimization-based~\cite{Mukadam_2018} and sampling-based~\cite{10610192} planning, and vectorized sampling~\cite{10611190}.

\medskip
\subsection{Learning-based mobile manipulation.}
Modular systems ground language instructions into learned environment representations or pre-trained skills~\cite{ahn2022do, huang2023voxposer, liu2024okrobot, yenamandra2024homerobotopenvocabularymobilemanipulation}.
Reinforcement learning trains end-to-end or modular whole-body policies at scale~\cite{jauhri2022robot, zhang2024gammagraspabilityawaremobilemanipulation}.
Vision-language-action and embodied foundation models~\cite{intelligence2025pi05visionlanguageactionmodelopenworld, tu2026sgvlalearningspatiallygroundedvisionlanguageaction, wu2025momanipvlatransferringvisionlanguageactionmodels, zitkovich2023rt} add semantic breadth but degrade on unseen layouts and objects.
These policies act without explicit geometric reasoning about collision risk and remain sensitive to distributional shift.
Our system offers a complementary model-based alternative that reasons explicitly about visibility and swept volumes within a closed loop coupling perception and planning in dynamic, unstructured environments.

%% file: formulation/formulation.tex
\begin{figure*}[t]
    \centering
    \resizebox{\textwidth}{!}{\input{figures/system_architecture}}
    \caption{\textbf{System Architecture Overview.}
        The receding-horizon framework updates the belief summary $b_t = (q_t, M_t, g_t)$ through four coupled components. The active perception policy directs gaze while updating the map $M_t$, the hierarchical subgoal policy generates goals $g_t$, the whole-body motion policy computes collision-free trajectories $\xi_t$ to $g_t$, and localization and control executes the trajectory while tracking state $q_t$. New observations can invalidate both trajectories and goals, triggering replanning or goal resampling.}
    \label{fig:system_overview}
\end{figure*}

\section{Problem Formulation}
\label{sec:formulation}

We consider mobile grasping in unknown, dynamic environments using a single RGB-D head camera with limited field of view.
The robot configuration is $q_t=(q_{r,t},q_{v,t})\in\mathcal{C}_r\times\mathcal{C}_v$, comprising the base, torso, and arm ($q_{r,t}$) and pan-tilt head ($q_{v,t}$).
Given $q_0$ and approximate target location $p$ in workspace $\mathcal{W}\subset\mathbb{R}^3$, we seek a stable grasp, maximizing success probability with short, collision-free execution.
The target is static until grasped, while surrounding obstacles follow unknown trajectories.

Let $M_t^*$ denote the true scene geometry and $M_t$ the map maintained from partial observations.
Observations describe the scene only at acquisition. Previously observed free space can become occupied, and visible targets can become occluded.
Thus, a trajectory validated in $M_t$ can become unsafe during execution.
The robot maintains a compact belief summary $b_t=(q_t,M_t,g_t)$, where $g_t\in\mathcal{C}_r$ is the current grasp or intermediate subgoal.

The two visibility constraints in Figure~\ref{fig:visibility_factor_graph} couple the subtasks of \emph{whole-body motion planning} and \emph{inverse kinematics (IK) for goal selection}.

The \emph{collision visibility constraint} $\phi_{colvis}$ requires observation coverage of the whole-body swept volume to be refreshed before the robot enters it.
Motion planning seeks a kinematically feasible trajectory $\xi_t:[0,H_t]\to\mathcal{C}_r$ from $q_{r,t}$ to $g_t$, collision-free in $M_t$, where $H_t$ is its duration.
Its swept volume $V(\xi_t)\subset\mathcal{W}$ is the union of robot volumes along it.
We address $\phi_{colvis}$ through \emph{active perception}, choosing admissible pan-tilt commands $a_v\in\mathcal{A}_v$ to maximize weighted visibility of $V(\xi_t)$, prioritizing imminent and fast-moving regions, including previously observed ones.

The \emph{object visibility constraint} $\phi_{objvis}$ requires the goal to admit a head orientation $v\in\mathcal{C}_v$ that keeps the target visible for grasp refinement, respecting camera limits and robot and mapped occlusions.
This becomes a \emph{visibility-constrained IK} subtask. For a candidate grasp pose $T_{ee}\in\mathrm{SE}(3)$ inferred from observations, we seek a collision-free $g_t$ in $M_t$ satisfying $F_{ee}(g_t)=T_{ee}$ and $\phi_{objvis}(g_t;M_t,p)=1$, where $F_{ee}$ is the end-effector forward kinematics.
The hierarchical subgoal policy samples and verifies candidates, falling back to pre-grasp or observation goals with target visibility when direct grasping fails.

Both constraints compete for the camera.
The receding-horizon policy $\pi=(\pi_v,\pi_g,\pi_r)$ couples active perception, subgoal selection, and motion planning using $b_t$ and target $p$.
Observations refresh $M_t$ during execution, triggering replanning or goal resampling as needed.

%% file: figures/system_architecture.tex
\usetikzlibrary{arrows.meta}%
\begingroup
\definecolor{archTeal}{HTML}{287D7A}%
\definecolor{archBlue}{HTML}{356EA0}%
\definecolor{archPurple}{HTML}{7761A3}%
\definecolor{archCopper}{HTML}{A56F3D}%
\definecolor{archAmber}{HTML}{B68A35}%
\definecolor{archInk}{HTML}{28343E}%

\tikzset{
    pics/gripper/.style={code={
        \draw (0,0.16) -- (0,0.06);
        \draw (-0.14,-0.14) -- (-0.14,0.06) -- (0.14,0.06)
            -- (0.14,-0.14);
        \draw (-0.14,-0.14) -- (-0.07,-0.14);
        \draw (0.14,-0.14) -- (0.07,-0.14);
    }},
    pics/arm/.style={code={
        \draw (-0.16,-0.14) -- (-0.04,-0.14) -- (-0.04,-0.08);
        \draw (-0.04,-0.08) -- (-0.12,0.08) -- (0.06,0.15)
            -- (0.15,0.02);
        \draw (-0.04,-0.08) circle (0.035);
        \draw (-0.12,0.08) circle (0.035);
        \draw (0.06,0.15) circle (0.035);
    }},
    pics/camera/.style={code={
        \draw[rounded corners=0.5pt] (-0.17,-0.12) rectangle (0.17,0.10);
        \draw (-0.09,0.10) -- (-0.06,0.15) -- (0.06,0.15) -- (0.09,0.10);
        \draw (0,-0.01) circle (0.065);
    }},
    pics/simplify/.style={code={
        \draw[opacity=0.4] (-0.15,-0.12) -- (-0.07,0.07)
            -- (0.05,-0.02) -- (0.15,0.13);
        \draw (-0.15,-0.12) -- (0.15,0.13);
        \fill (-0.15,-0.12) circle (0.035);
        \fill (0.15,0.13) circle (0.035);
    }},
    pics/route/.style={code={
        \draw[rounded corners=1pt] (-0.14,-0.12) -- (-0.14,0.10)
            -- (0.02,0.10) -- (0.02,-0.02) -- (0.15,-0.02)
            -- (0.15,0.14);
        \fill (-0.14,-0.12) circle (0.035);
        \fill (0.15,0.14) circle (0.035);
    }},
    pics/locate/.style={code={
        \draw (0,0) circle (0.115);
        \draw (-0.18,0) -- (-0.08,0);
        \draw (0.08,0) -- (0.18,0);
        \draw (0,-0.18) -- (0,-0.08);
        \draw (0,0.08) -- (0,0.18);
        \fill (0,0) circle (0.025);
    }},
    pics/control/.style={code={
        \draw (-0.13,-0.16) -- (-0.13,0.16);
        \draw (0,-0.16) -- (0,0.16);
        \draw (0.13,-0.16) -- (0.13,0.16);
        \draw[fill=white] (-0.17,0.02) rectangle (-0.09,0.09);
        \draw[fill=white] (-0.04,-0.10) rectangle (0.04,-0.03);
        \draw[fill=white] (0.09,0.04) rectangle (0.17,0.11);
    }},
    pics/eye/.style={code={
        \draw (-0.18,0) .. controls (-0.06,0.17) and (0.06,0.17)
            .. (0.18,0) .. controls (0.06,-0.17) and (-0.06,-0.17)
            .. cycle;
        \draw (0,0) circle (0.06);
        \fill (0,0) circle (0.02);
    }},
    pics/map/.style={code={
        \draw (-0.16,-0.13) -- (-0.055,-0.17) -- (0.055,-0.11)
            -- (0.16,-0.15) -- (0.16,0.13) -- (0.055,0.17)
            -- (-0.055,0.11) -- (-0.16,0.15) -- cycle;
        \draw (-0.055,-0.17) -- (-0.055,0.11);
        \draw (0.055,-0.11) -- (0.055,0.17);
    }}
}%
\begin{tikzpicture}[
    x=1.35cm, y=-1cm,
    font=\fontsize{11.5}{13}\selectfont,
    text=archInk,
    block/.style={draw=archInk, fill=white, line width=0.85pt,
        rectangle, rounded corners=2pt,
        align=center, inner sep=2pt, outer sep=0pt},
    subgoal/.style={block, draw=archTeal!80!black},
    motion/.style={block, draw=archBlue!80!black},
    perception/.style={block, draw=archPurple!80!black},
    control/.style={block, draw=archCopper!80!black, fill=archCopper!6},
    frame/.style={line width=0.85pt, rounded corners=7pt},
    title/.style={font=\fontsize{12}{14}\selectfont\bfseries, inner sep=0pt},
    note/.style={text=black!65, font=\fontsize{10.5}{12}\selectfont,
        align=center, inner sep=1pt},
    wire/.style={draw=black!60, line width=0.85pt},
    flow/.style={wire, -{Latex[length=2.3mm,width=1.8mm]}}
]
    \path[use as bounding box] (0,0) rectangle (20.48,6.86);

    \draw[frame, draw=archTeal!70, fill=archTeal!6]
        (2.50,0.14) rectangle (7.55,4.80);
    \draw[frame, draw=archBlue!70, fill=archBlue!6]
        (7.94,0.14) rectangle (12.05,4.80);
    \draw[frame, draw=archPurple!70, fill=archPurple!6]
        (16.14,0.14) rectangle (20.34,4.80);
    \node[title, text=archTeal!85!black] at (5.025,0.43) {Subgoal Generation};
    \node[title, text=archBlue!85!black] at (9.995,0.43) {Motion Planner};
    \node[title, text=archPurple!85!black] at (18.24,0.43) {Perception};

    \node[subgoal, minimum width=4.5225cm, minimum height=0.83cm]
        (grasp) at (5.02,1.26) {Grasping pose};
    \node[subgoal, minimum width=4.5225cm, minimum height=0.83cm]
        (pregrasp) at (5.02,2.64) {Pre-pose};
    \node[subgoal, minimum width=4.5225cm, minimum height=0.83cm]
        (observe) at (5.02,4.02) {Observation Pose};
    \node[motion, minimum width=4.3875cm, minimum height=0.83cm]
        (simplify) at (9.97,1.32) {Simplification};
    \node[motion, minimum width=4.3875cm, minimum height=0.83cm]
        (planner) at (9.97,2.64) {Vectorized planner};

    \node[control, minimum width=3.4425cm, minimum height=1.10cm]
        (localize) at (14.57,2.05) {Localization\\System};
    \node[control, minimum width=4.536cm, minimum height=1.33cm]
        (controller) at (14.14,4.09) {MPC Whole-body\\Tracking Controller};
    \node[perception, minimum width=4.5225cm, minimum height=1.69cm]
        (perception) at (18.225,1.835) {Active Perception};
    \node[perception, minimum width=4.5225cm, minimum height=0.83cm]
        (map) at (18.245,3.82) {Map Update};

    \foreach \component/\archglyph/\accent in {
        grasp/gripper/archTeal,
        pregrasp/arm/archTeal,
        observe/camera/archTeal,
        simplify/simplify/archBlue,
        planner/route/archBlue,
        localize/locate/archCopper,
        controller/control/archCopper,
        perception/eye/archPurple,
        map/map/archPurple%
    } {
        \begin{scope}[shift={(\component.west)}, xshift=0.31cm,
            x=1cm, y=1cm, color=\accent, line width=0.75pt,
            line cap=round, line join=round]
            \pic {\archglyph};
        \end{scope}
    }

    \draw[draw=archAmber!85!black, fill=archAmber!9, line width=0.85pt]
        (11.80,5.93) -- (14.14,5.15)
        -- (16.46,5.93) -- (14.14,6.72) -- cycle;
    \node[align=center, inner sep=0pt] at (14.14,5.93)
        {Imminent\\collision?};

    \draw[flow] (0.07,1.19) -- (3.345,1.19);
    \node[note] at (1.25,0.66) {Rough Target\\Object Position};
    \draw[flow] (3.99,1.675) -- (3.99,2.225);
    \node[note, anchor=west] at (4.03,1.95) {Not detected};
    \draw[flow] (3.99,3.055) -- (3.99,3.605);
    \node[note, anchor=west] at (4.03,3.33) {Not detected};

    \draw[wire] (grasp.east) -- (7.12,1.26) -- (7.12,4.02)
        -- (observe.east);
    \draw[flow] (pregrasp.east) -- (planner.west);
    \draw[flow] (planner.north) -- (simplify.south);

    \draw[flow] (simplify.east) -- (16.55,1.32);
    \node[note] at (12.96,1.08) {Waypoints};
    \draw[flow] (12.96,1.32) -- (12.96,3.425);
    \draw[flow] (14.49,3.425) -- (14.49,2.60);
    \draw[flow] (localize.east) -- (16.55,2.05);
    \draw[flow] (14.02,1.05) -- (16.55,1.05);
    \node[note] at (14.95,0.59) {Rough Target\\Object Position};
    \draw[flow] (perception.south) -- (map.north);

    \draw[wire] (17.96,4.235) -- (17.96,5.93) -- (16.46,5.93);
    \node[note] at (17.16,5.70) {[10Hz]};
    \draw[flow] (14.14,5.15) -- (controller.south);
    \node[note, anchor=west] at (14.25,5.04) {No};
    \draw[wire] (11.80,5.93) -- (2.05,5.93) -- (2.05,1.19);
    \node[note] at (6.51,5.69) {Goal collision};
    \draw[flow] (9.97,5.93) -- (planner.south);
    \node[note, anchor=west, align=left] at (10.09,5.38)
        {Waypoint\\collision};
    \draw[flow] (18.33,5.98) -- (18.33,4.235);
    \node[note] at (19.30,5.42) {Canonical\\Map (Partial\\or empty)};
\vspace{-0.7cm}
\end{tikzpicture}%
\vspace{-0.7cm}
\endgroup%

%% file: method/method.tex
\section{Method}
\label{sec:method}

Our system realizes the policy $\pi = (\pi_{v}, \pi_{g}, \pi_{r})$ defined in Section~\ref{sec:formulation} through a receding-horizon framework (Figure~\ref{fig:system_overview}).
In each cycle, the robot updates the map $M_t$ with new observations and checks whether the current subgoal and trajectory remain valid.
Throughout the cycle, the \textbf{active perception policy} $\pi_{v}$ directs the gaze to the current bottleneck of the task, the target region before grasping and the trajectory's swept volume during motion (\secref{subsec:partial_observation}).
When the subgoal or trajectory is invalidated or a subgoal is reached, the \textbf{subgoal policy} $\pi_{g}$ proposes a new subgoal $g_t$ through a behavior tree that escalates from direct grasping to pre-grasp repositioning and observation gathering (\secref{subsec:subgoal_generation}).
The \textbf{whole-body motion policy} $\pi_{r}$ then replans a collision-free trajectory $\xi_t$ fast enough to keep pace with map changes (\secref{subsec:motion_planning}).
This coupling refreshes observations of the space the robot is about to sweep, allowing replanning when dynamic obstacles invalidate earlier observations.

\subsection{Active Perception Policy \texorpdfstring{$\pi_{v}$}{πv}}
\label{subsec:partial_observation}
\input{method/partial_observation}

\subsection{Subgoal Policy \texorpdfstring{$\pi_{g}$}{πg}}
\label{subsec:subgoal_generation}
\input{method/subgoal_generation}

\begin{figure*}[t]
    \centering
    \includegraphics[width=0.95\linewidth]{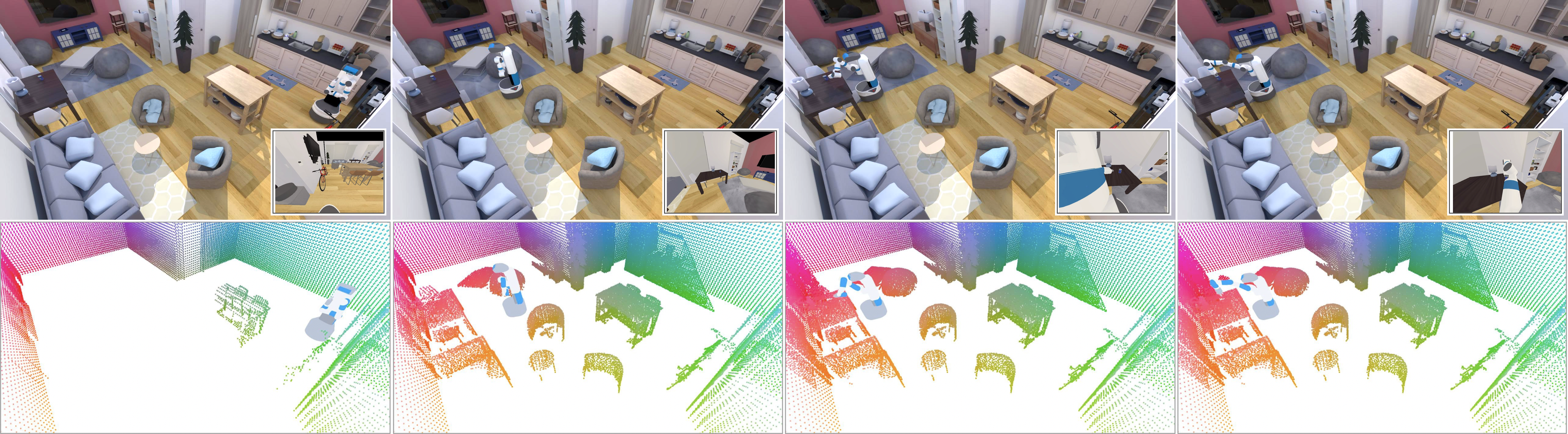}
    \caption{Qualitative results showing the execution sequence in simulation. From left to right: starting pose, observation stage, pre-grasp stage, and grasping stage. The top row shows the scene with the head camera view inset, and the bottom row shows the robot's belief map.}
    \label{fig:execution_sequence}
    \vspace{-0.4cm}
\end{figure*}

\subsection{Whole-Body Motion Policy \texorpdfstring{$\pi_{r}$}{πr}}
\label{subsec:motion_planning}
\input{method/motion_planning}

%% file: method/partial_observation.tex
\begin{figure}[t]
    \centering
    \includegraphics[width=\columnwidth]{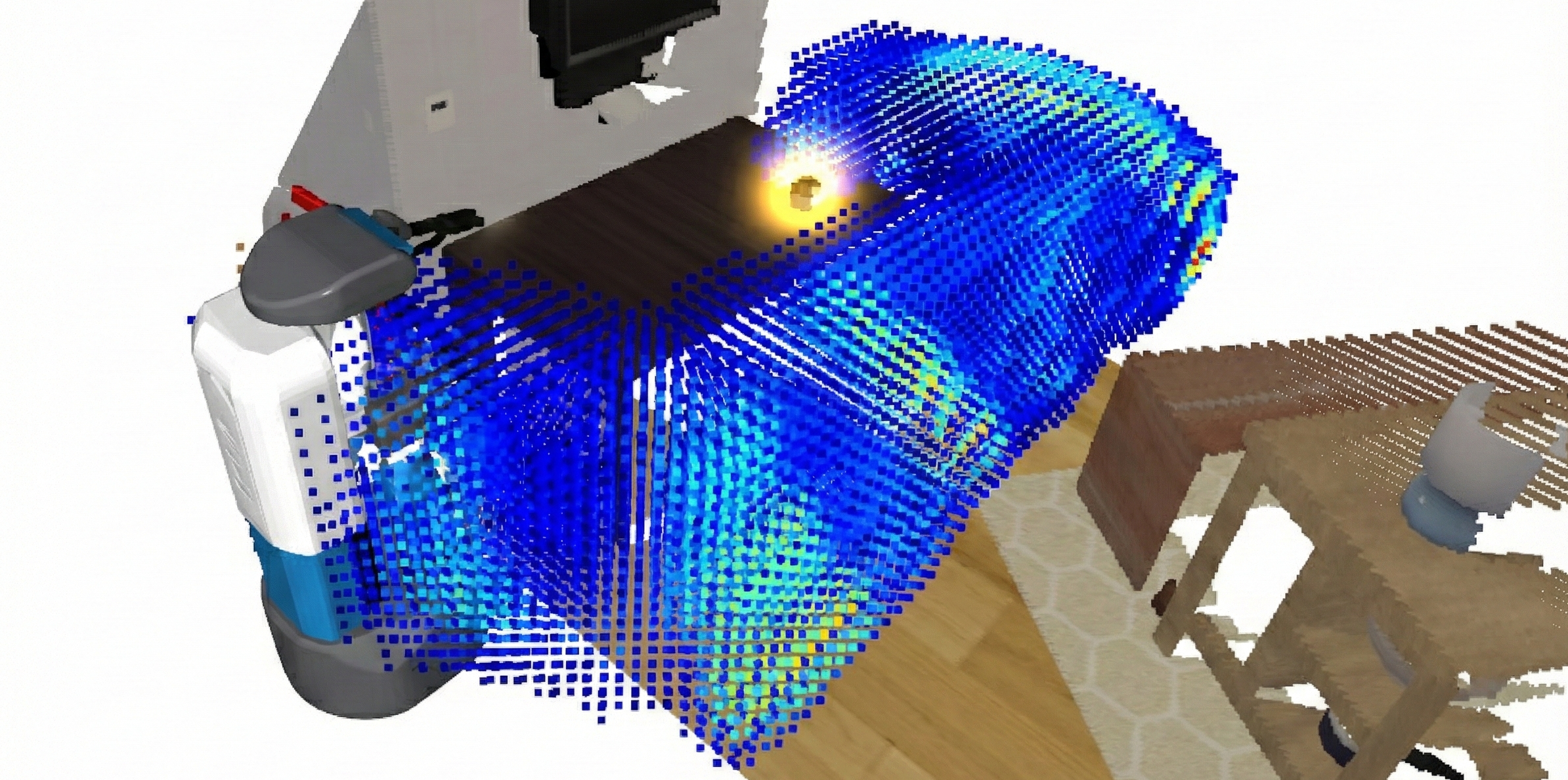}
    \caption{\textbf{State-Dependent Gaze Optimization.}
        During planning (\(\xi_t = \emptyset\)), the importance field concentrates on the target region to enable grasp pose generation. During execution (\(\xi_t \neq \emptyset\)), the field shifts to the swept volume \(V(\xi_t)\) (color-coded heatmap), weighted by velocity and temporal proximity to prioritize observation of collision-critical regions.}
    \label{fig:active_perception}
\vspace{-0.5cm}
\end{figure}

A robot with a limited field of view must balance two competing demands, observing the target \(p\) for stable grasping and monitoring the swept volume \(V(\xi_t)\) for collision safety.
These demands are temporally separable.
Target observation matters most during planning, while swept-volume monitoring matters during execution.
We formulate gaze control as maximizing visibility of an importance field \(\Phi(x)\) that adapts to the robot's phase:
\begin{equation}
    a_v^* = \operatorname*{argmax}_{a_v \in \mathcal{A}_v} \sum_{x \in \mathcal{W}} \Phi(x) \cdot \mathcal{V}(x, a_v)
\end{equation}
where \(\mathcal{V}(x, a_v) \in \{0,1\}\) indicates visibility under gaze \(a_v\).
When no trajectory is being executed, the gaze observes the target area:
\begin{equation}
    \Phi(x) =\mathbb{I}(x \in \mathcal{B}_\epsilon(p))
\end{equation}
where \(\mathcal{B}_\epsilon(p)\) is a ball around the target to support grasp generation.
Once a trajectory \(\xi_t\) is being executed, the field shifts to collision-critical regions along it:
\begin{equation}
\Phi(x) =\mathbb{I}(x \in V(\xi_t)) \cdot w_{safety}(x)
\end{equation}
where \(w_{safety}(x)\) focuses on near-future and high-velocity regions (Figure~\ref{fig:active_perception}):
\begin{equation}
    w_{safety}(x) = \sum_{i=0}^{N} \gamma^i \cdot w_d(x, q^{(i)}) \cdot \|\dot{q}^{(i)}\|
    \label{eq:velocity-weights}
\end{equation}
where \(i\) indexes a waypoint in \(\xi_t\), \(\gamma^i\) decays with lookahead, \(\|\dot{q}^{(i)}\|\) is the velocity at waypoint \(i\), and \(w_d\) is a Gaussian centered at a moderate distance from the robot, where corrective action remains possible.
The formulation extends~\cite{finean2021where} with velocity-aware, time-discounted weighting of the swept volume.
We discretize the gaze space \(\mathcal{A}_v\) into a pan-tilt grid and evaluate the objective exhaustively.

We maintain the map \(M_t\) as a point cloud updated via ray casting, where new observations add perceived surface points as obstacles while removing previously occupied space along each ray. This enables the robot to detect when dynamic obstacles have moved, while preserving occluded regions. The update pipeline includes ground plane removal, robot self-filtering, and voxel-based point merging.

%% file: method/subgoal_generation.tex
Since the subgoal \(g_t\) depends on observed geometry, collision constraints, and robot reachability revealed as the map \(M_t\) changes, the subgoal policy \(\pi_g\) organizes grasping as a behavior tree with three fallback levels, namely grasp, pre-grasp, and observe goal sampling.
It attempts the most direct level first and escalates to a more exploratory one when the current level fails, which also allows recovery from failures at runtime.

The policy first attempts to grasp without base motion, maximizing efficiency when the target is already within reach.
We use a sample-then-verify strategy, in which a learning-based grasp network \cite{sundermeyer2021contact-graspnet} generates candidate end-effector poses from the observed point cloud.
These candidates are filtered by a pre-computed capability map~\cite{9517662}. We select \(N\) diverse poses via farthest point sampling and rank them by grasp confidence with a small penalty on the distance to the current end-effector pose. Each candidate is validated for whole-body inverse kinematics \cite{beeson2015trac-ik} feasibility and collision against the current map \(M_t\), and the first valid configuration is selected as \(g_t\).
Execution reuses the receding-horizon loop, where the arm trajectory is monitored for collisions and replanned as the map updates, followed by a short linear approach before closing the gripper.

When direct grasping fails, the robot must resample whole-body configurations that balance visibility and manipulability.
Rather than solving a computationally expensive whole-body inverse kinematics problem, we factor the search over base pose, torso height, and end-effector pose into three conditional sampling distributions.
The distribution \(p_{geo}\) samples collision-free base poses \(q_b\) from a 2D signed distance field that is updated in real-time from the point cloud.
Given the sampled base pose, \(p_{torso}\) queries a pre-built torso height map to select \(h\) that maximizes the end-effector's reach to target \(p\).
Finally, \(p_{ee}\) samples end-effector poses \(T_{ee}\) in a sphere around the target, filtered by the capability map.
Each sampled whole-body configuration \((q_b, h, T_{ee})\) is then validated via rejection sampling for inverse kinematics feasibility, collision, and self-occlusion, where the last check preserves visibility of the target.
Both the capability map and the torso height map are pre-computed offline from large-scale sampled forward kinematics.
The former inverts a reachability map to test whether an end-effector pose is reachable relative to the base, and the latter stores, for each end-effector position, the torso height that maximizes the number of feasible arm configurations.

When manipulation is blocked by occlusion or unknown geometry, the policy falls back to gathering observations that update the map \(M_t\). We sample \(N\) base poses at a fixed safe standoff distance from the target, each paired with a robust arm configuration.
These poses maintain visibility of the target \(p\) while minimizing displacement from the current body configuration \(q_{r,t}\). If all three strategies fail, the policy signals failure to the behavior tree, triggering task termination.

%% file: method/motion_planning.tex
Because $M_t$ changes during execution, $\pi_r$ must rapidly replan smooth, collision-free whole-body trajectories. Existing planners face a trade-off between speed and generality. VAMP~\cite{10611190} achieves real-time performance but assumes homogeneous configuration spaces, whereas non-holonomic mobile-manipulator planners~\cite{10610192} support heterogeneous base-arm spaces at higher computational cost. We combine their strengths to achieve both real-time performance and kinematic generality.

The key challenge is extending vectorized collision checking to non-holonomic steering. We therefore decompose planning hierarchically following~\cite{10610192}. The algorithm samples base placements using a non-holonomic kernel with Hybrid A* \cite{Kurzer1057261} for global guidance and Reeds-Shepp curves for local connections, then connects the corresponding arm configurations with RRT-Connect \cite{844730}. To respect their distinct kinematics, base and arm motions are discretized with Reeds-Shepp curves and linear interpolation, respectively, then synchronized at matching time steps. The resulting whole-body configurations support parallel SIMD validation \cite{10611190}, retaining vectorization speed without violating either kinematic structure.

After planning, randomized shortcutting~\cite{Geraerts2007, Hauser2010} removes unnecessary waypoints, B-spline fitting~\cite{Pan2012} smooths the trajectory, and dense re-interpolation validates it for collision-free execution. The resulting pipeline enables responsive replanning as $M_t$ changes.

A whole-body velocity tracker executes the planned trajectory.
At 20\,Hz, it linearizes the non-holonomic base dynamics around the current heading and solves a one-step quadratic program in closed form, mapping the 11-D configuration error to a 10-D velocity command for the base, torso, and arm.
This joint optimization keeps the three synchronized, which is critical because uncoordinated motion causes the end-effector to deviate.
The tracker accepts trajectory updates during execution, enabling reactive replanning.

%% file: experiments/experiments.tex
\section{Experiments}
\label{sec:experiments}

We evaluated our system across randomized simulation scenarios and real-world trials.
Our experiments address three questions.
First, how much does closing the loop between perception, subgoal reasoning, and whole-body planning improve over the prevailing decoupled designs?
Second, how much of the improvement comes from velocity-aware active perception, and through which failure modes?
Third, what structural limitations remain even with the full system?

\input{experiments/experimental_setup}

\input{experiments/env_setup}
\input{experiments/metrics}
\input{experiments/baselines}

%% file: experiments/experimental_setup.tex
\subsection{Experimental Setup}
\label{subsec:experimental_setup}

\paragraph{Benchmark}
For simulation experiments, we used ManiSkill3~\cite{taomaniskill3} with ReplicaCAD scenes~\cite{szot2021habitat} for photorealistic rendering and GPU-accelerated physics, modeling a Fetch mobile manipulator with accurate kinematics and dynamics.
For real-world experiments, we used a Fetch mobile manipulator with ROS and localized the base with AMCL.

%% file: experiments/env_setup.tex
We evaluated three complementary scenarios. \textit{Unknown static environments} tested perception and planning under visibility constraints with previously unseen layouts. \textit{Dynamic environments} evaluated online perception and replanning against a moving obstacle. \textit{Real robot deployment} validated real-world performance with localization and model errors.
All scenarios were room scale (\(\sim\)5-10m navigation range) and contained at least one collision-free path.

We generated 400 test scenarios (20 scenes $\times$ 20 objects) with procedural YCB object placement following~\cite{zhang2024gammagraspabilityawaremobilemanipulation}. Each placement underwent validation:
A drop-check verified object stability.
An oracle grasp-check validated grasp feasibility using physics simulation following ACRONYM~\cite{eppner2021acronym}.
For each dynamic scenario, a collision-enabled obstacle with a $0.10 \times 0.20$\,m footprint and 1.75\,m height was instantiated on the active navigation route approximately 1.45\,m ahead of the robot and moved transversely at 0.12\,m/s throughout execution. The robot received no oracle obstacle state or trajectory. We accepted a placement only when the navigation mesh retained a collision-free detour to the target, and counted the dynamic interaction as valid only when replanning measurably changed the remaining route. Over the dynamic runs, placement succeeded in 97.5\% of trials, replanning changed the route in 92.1\%, and the mean closest approach to the obstacle was 0.81\,m.
For the learning baselines, we report Easy and Hard evaluation protocols. Easy follows the MS-HAB Pick initialization by placing the robot 0.6--1.6~m from and facing the target with a 200-step horizon. Hard uses exactly the same 400 tasks, robot starts, moving-obstacle settings, and physical success criterion as our benchmark with a 1{,}000-step horizon.

We tested at five representative indoor locations (dining table, kitchen counter, workstation, coffee table, sofa) with 40 configurations total: 20 in static unknown and 20 in dynamic scenarios with movable objects.

%% file: experiments/metrics.tex
\paragraph{Evaluation Metrics} We evaluated system performance using grasp success rate as the primary metric.
A trial succeeded if the robot lifted the target object above a height threshold (10\,cm) and maintained a stable grasp, which means the object remained stable in the gripper for more than 2 seconds.
We additionally report success weighted by path length (SPL), the ratio of the shortest collision-free path to the executed path over successful trials, to measure path efficiency.
We categorized failures into three groups: \textit{execution failures} (collision, grasp, IK) captured errors during physical interaction, \textit{system limitations} (reachability, perception) reflected hardware and sensing constraints, and \textit{planning failures} indicated that no valid subgoal or collision-free path was found within the planner's budget of three 200\,ms attempts.

%% file: experiments/baselines.tex
\paragraph{Baselines and Ablations} We compare our full system against seven baselines and ablations spanning learning-based policies, conventional mobile manipulation designs, and controlled ablations.

\noindent\textbf{SG-VLA}
SG-VLA~\cite{tu2026sgvlalearningspatiallygroundedvisionlanguageaction} is a vision-language-action policy for mobile manipulation.
As its original code is unavailable, we reimplemented its spatially grounded visual representation and trained it on Pick demonstrations from the MS-HAB pipeline.

\noindent\textbf{RL (MS-HAB)}
This baseline uses the MS-HAB SAC Pick policy with its original training pipeline and configuration, directly predicting continuous whole-body actions from visual and robot-state observations.

\noindent\textbf{Direct Grasping}
This baseline executes a fixed observation, pre-grasp, and grasp sequence with whole-body replanning, removing adaptive subgoal resampling and structured recovery to isolate the benefit of the subgoal policy \(\pi_{g}\) and behavior tree.

\noindent\textbf{CapMap Placement}
This baseline selects a kinematically feasible base pose with our pre-grasp sampler and capability-map filtering~\cite{9830863}, navigates there with the arm tucked, then switches to fixed-base arm planning.

\noindent\textbf{Navigation-and-manipulation}
Following the dominant decoupled paradigm~\cite{9830863, 6630839}, the robot navigates with its arm tucked to a base pose at a fixed target distance, then switches to fixed-base arm planning, replanning navigation after collisions.

\noindent\textbf{Ours (w/o Velocity Weighting)}
This ablation removes velocity weighting from the gaze objective in Eq. \ref{eq:velocity-weights}, keeping only volume-based safety weights~\cite{finean2021where}, to isolate velocity-aware active perception.

\noindent\textbf{Ours (w/o Observation Stage)}
This ablation removes the dedicated observation stage and its structured behavior transitions, repeatedly replanning toward sampled pre-grasp configurations, to isolate the contribution of perception behavior and high-level task management.

\begin{figure}[t]
    \centering
    \includegraphics[width=0.9\linewidth]{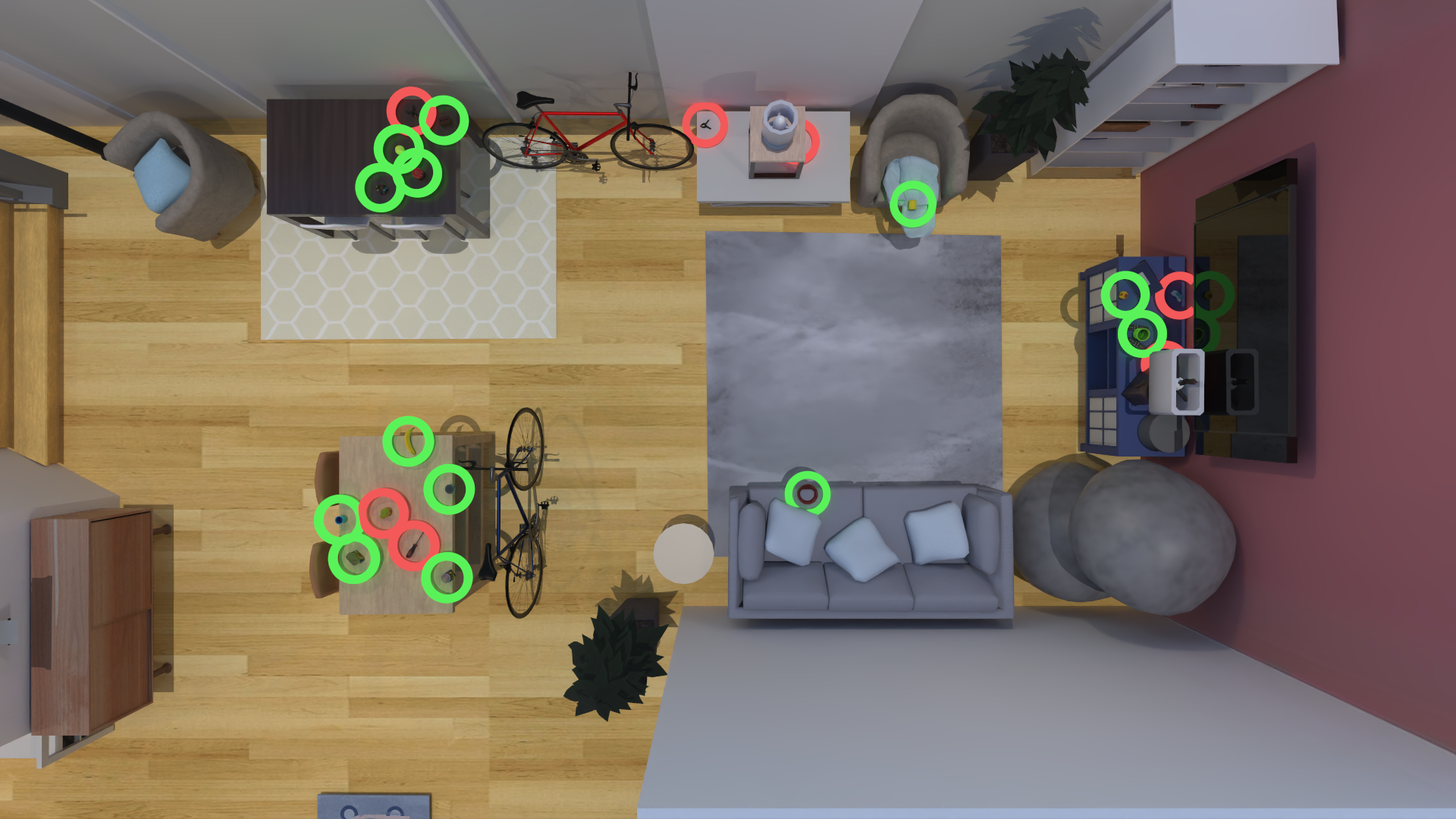}
    \caption{\textbf{Spatial Distribution of Success and Failure.} Green and red circles represent successful and failed grasping attempts, respectively.}
    \label{fig:success_distribution}
    \vspace{-0.4cm}
\end{figure}

%% file: results/results.tex
\input{results/simulation_results}

%% file: results/simulation_results.tex
\subsection{Results and Analysis}
\label{subsec:simulation_results}

\begin{table*}[t]
    \centering
    \caption{Simulation success rates and failure breakdowns in percent. Entries report the mean with the 95\% confidence-interval half-width from Student's $t$ distribution over five runs as a subscript. Lower is better for all failure categories. For RL and SG-VLA, grasp failures combine no-contact and unstable-hold outcomes. \emph{Inval.} denotes dynamic trials without a measurable route change and applies only to the dynamic condition. Dashes indicate unavailable or inapplicable categories.}
    \label{tab:results_sim}
    \scriptsize
    \setlength{\tabcolsep}{2.2pt}
    \renewcommand{\arraystretch}{0.96}
    \begin{tabular}{@{}lccccccccccccccc@{}}
        \toprule
        & \multicolumn{7}{c}{\textbf{Unknown Static Environment}} & \multicolumn{8}{c}{\textbf{Unknown Dynamic Environment}} \\
        \cmidrule(lr){2-8} \cmidrule(l){9-16}
        \textbf{Method} & \textbf{Succ.} $\uparrow$ & \textbf{Coll.} & \textbf{Grasp} & \textbf{IK} & \textbf{Reach.} & \textbf{Perc.} & \textbf{Plan.} & \textbf{Succ.} $\uparrow$ & \textbf{Coll.} & \textbf{Grasp} & \textbf{IK} & \textbf{Reach.} & \textbf{Perc.} & \textbf{Plan.} & \textbf{Inval.} \\
        \midrule
        SG-VLA (Easy)     & $1.6_{1.2}$ & $78.8_{1.9}$ & $19.6_{1.8}$ & -- & -- & -- & -- & $1.2_{0.8}$ & $80.5_{2.1}$ & $18.3_{1.9}$ & -- & -- & -- & -- & -- \\
        SG-VLA (Hard)     & $0.0_{0.0}$ & $94.8_{0.0}$ & $5.2_{0.0}$ & -- & -- & -- & -- & $0.0_{0.0}$ & $97.8_{0.0}$ & $2.2_{0.0}$ & -- & -- & -- & -- & -- \\
        RL (MS-HAB, Easy) & $26.9_{2.7}$ & $42.7_{3.1}$ & $30.4_{2.2}$ & -- & -- & -- & -- & $24.1_{3.0}$ & $44.9_{2.4}$ & $31.0_{2.8}$ & -- & -- & -- & -- & -- \\
        RL (MS-HAB, Hard) & $12.0_{0.0}$ & $80.8_{0.0}$ & $7.2_{0.0}$ & -- & -- & -- & -- & $9.5_{0.0}$ & $84.5_{0.0}$ & $6.0_{0.0}$ & -- & -- & -- & -- & -- \\
        Direct Grasping   & $27.2_{2.5}$ & $2.5_{0.8}$ & $1.6_{1.0}$ & $5.2_{0.8}$ & $2.5_{0.1}$ & $4.9_{1.2}$ & $56.3_{3.2}$ & $16.1_{1.1}$ & $14.8_{1.9}$ & $1.4_{0.5}$ & $1.4_{0.3}$ & $2.5_{0.0}$ & $4.0_{0.8}$ & $35.4_{1.6}$ & $24.4_{1.2}$ \\
        CapMap Placement  & $53.0_{2.4}$ & $9.7_{1.0}$ & $9.8_{2.1}$ & $0.0_{0.0}$ & $0.2_{0.0}$ & $7.5_{0.9}$ & $19.8_{1.3}$ & $49.7_{0.9}$ & $8.2_{0.8}$ & $9.1_{1.4}$ & $0.0_{0.0}$ & $0.0_{0.0}$ & $7.3_{1.0}$ & $16.3_{1.0}$ & $9.4_{0.3}$ \\
        Nav.-and-manip.   & $54.0_{2.4}$ & $7.7_{1.1}$ & $8.7_{1.1}$ & $0.0_{0.0}$ & $0.0_{0.0}$ & $7.6_{1.0}$ & $22.1_{1.5}$ & $48.6_{3.4}$ & $7.5_{1.7}$ & $8.7_{1.7}$ & $0.0_{0.0}$ & $0.0_{0.0}$ & $7.6_{1.3}$ & $18.1_{1.7}$ & $9.6_{0.5}$ \\
        w/o Vel.\ Weighting & $62.2_{1.8}$ & $12.3_{2.3}$ & $8.4_{0.6}$ & $0.3_{0.3}$ & $3.1_{0.3}$ & $3.0_{0.3}$ & $10.4_{0.3}$ & $51.5_{1.6}$ & $18.1_{2.4}$ & $7.0_{0.7}$ & $0.6_{0.3}$ & $3.0_{0.5}$ & $2.7_{0.1}$ & $8.8_{1.1}$ & $8.3_{1.2}$ \\
        w/o Obs.\ Stage   & $68.3_{3.7}$ & $5.8_{2.4}$ & $7.2_{2.9}$ & $0.2_{0.1}$ & $1.3_{0.3}$ & $2.6_{0.3}$ & $14.6_{1.3}$ & $56.3_{3.4}$ & $13.2_{4.4}$ & $6.7_{1.7}$ & $0.3_{0.3}$ & $2.2_{0.6}$ & $2.5_{1.1}$ & $11.3_{2.7}$ & $7.5_{1.0}$ \\
        Ours (Full System)       & $\mathbf{70.3}_{1.1}$ & $5.0_{1.5}$ & $8.0_{1.3}$ & $0.5_{0.3}$ & $3.0_{0.6}$ & $2.5_{0.9}$ & $10.7_{1.3}$ & $\mathbf{57.9}_{1.5}$ & $13.9_{3.4}$ & $5.7_{0.1}$ & $0.1_{0.1}$ & $2.1_{1.5}$ & $2.6_{0.6}$ & $10.7_{2.4}$ & $7.0_{1.2}$ \\
        \bottomrule
    \end{tabular}
\end{table*}

\begin{figure*}[t]
    \centering
    \includegraphics[width=0.245\linewidth,trim=0 90bp 0 70bp,clip]{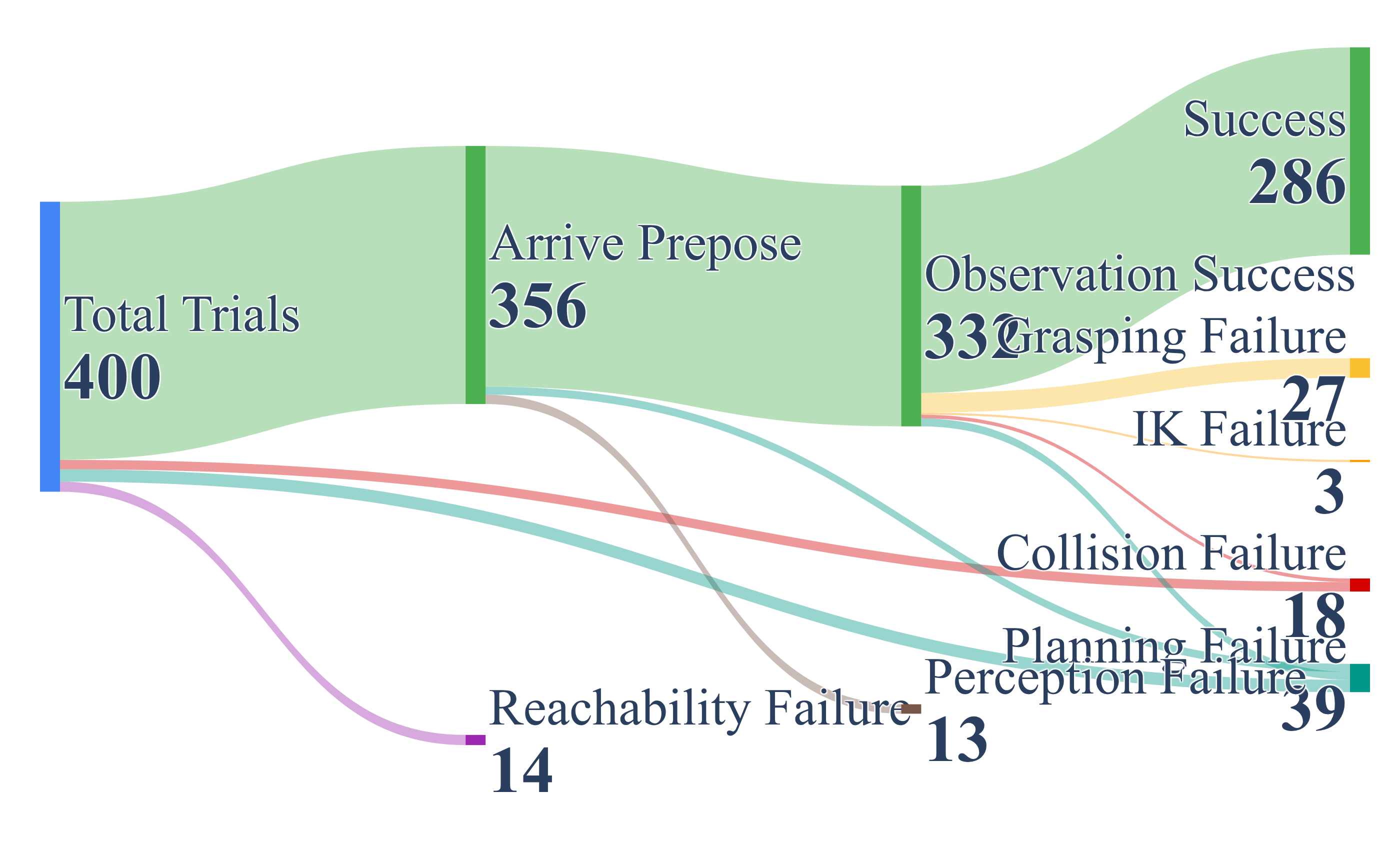}\hfill
    \includegraphics[width=0.245\linewidth,trim=0 90bp 0 70bp,clip]{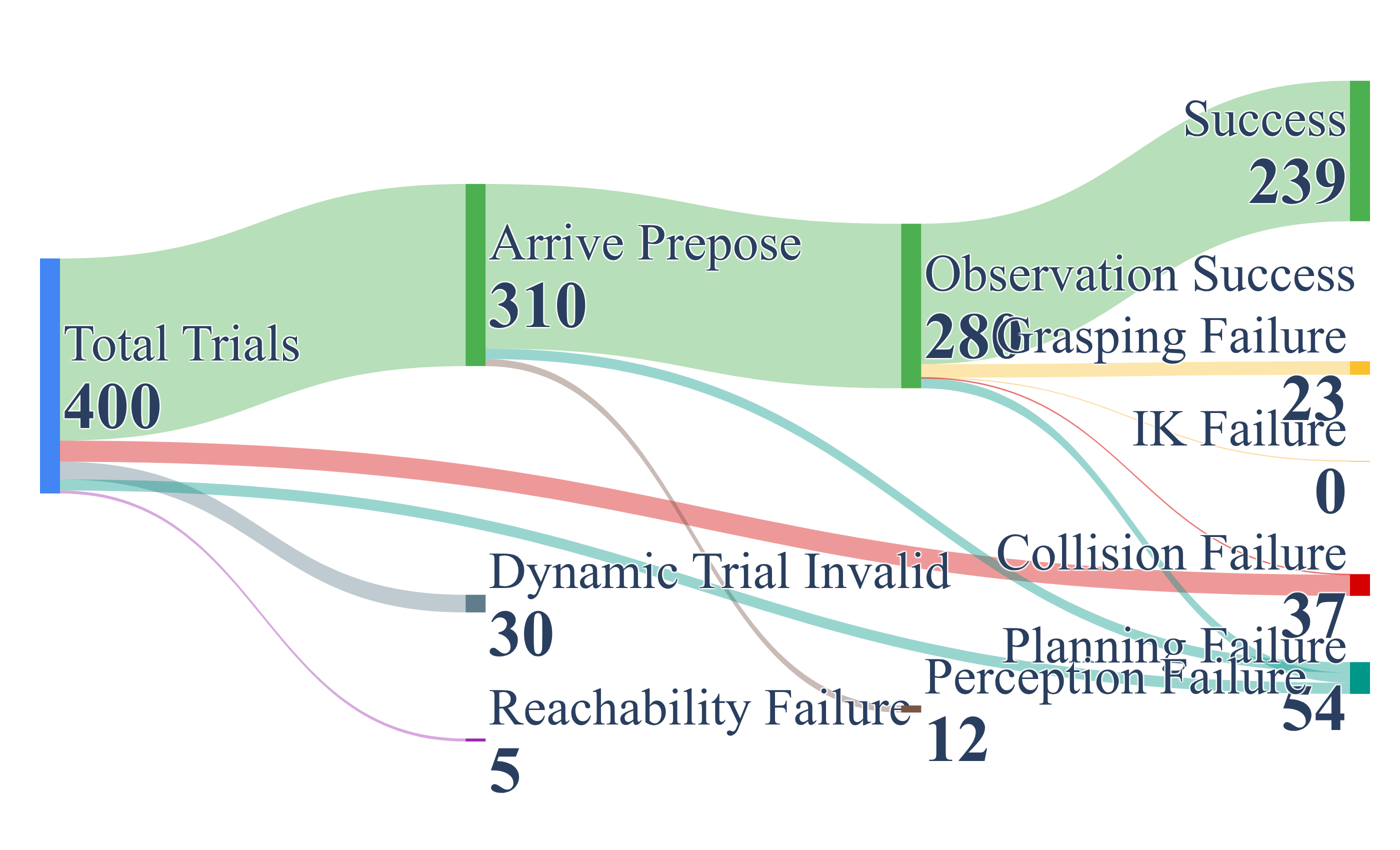}\hfill
    \includegraphics[width=0.245\linewidth,trim=0 90bp 0 70bp,clip]{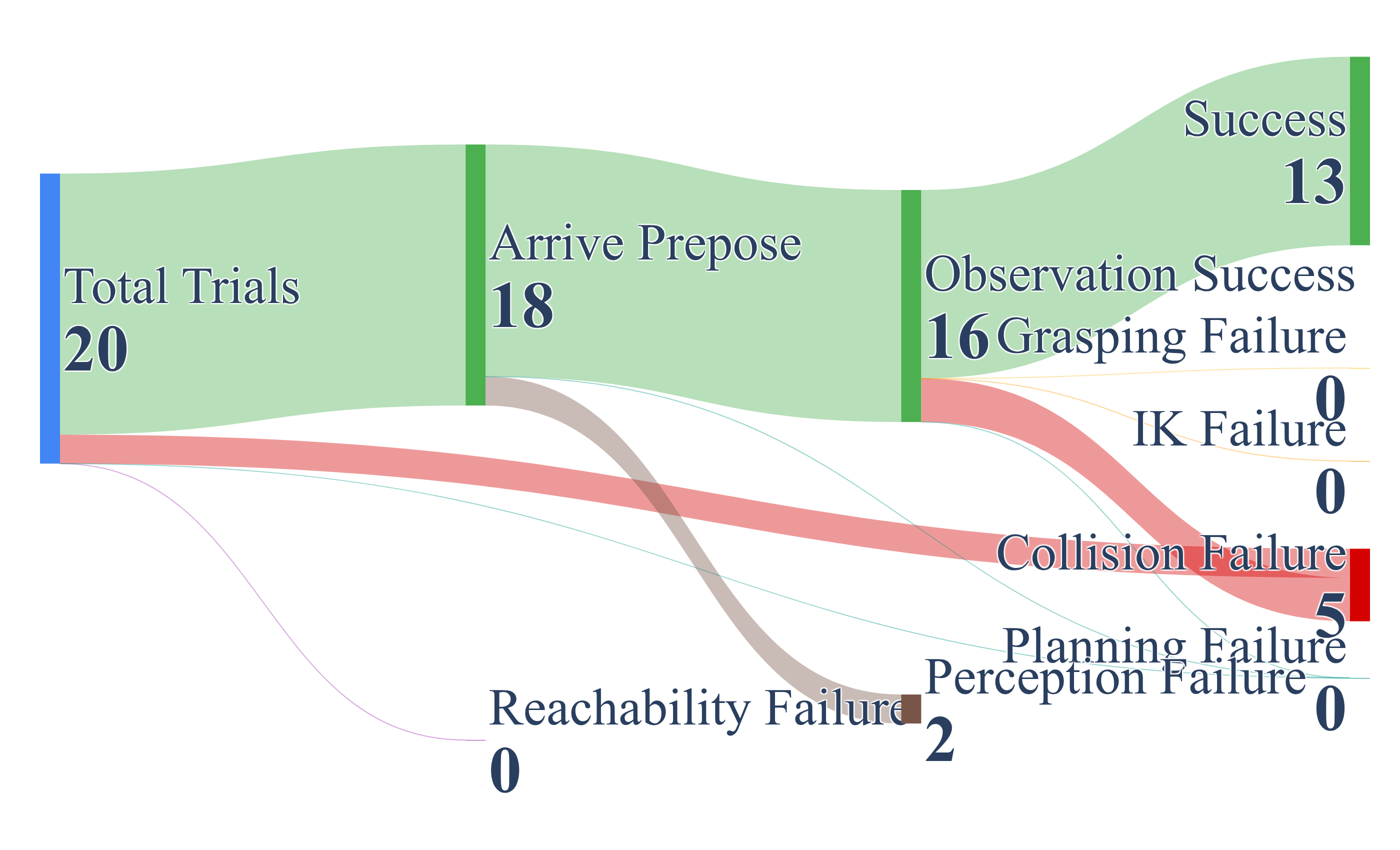}\hfill
    \includegraphics[width=0.245\linewidth,trim=0 90bp 0 70bp,clip]{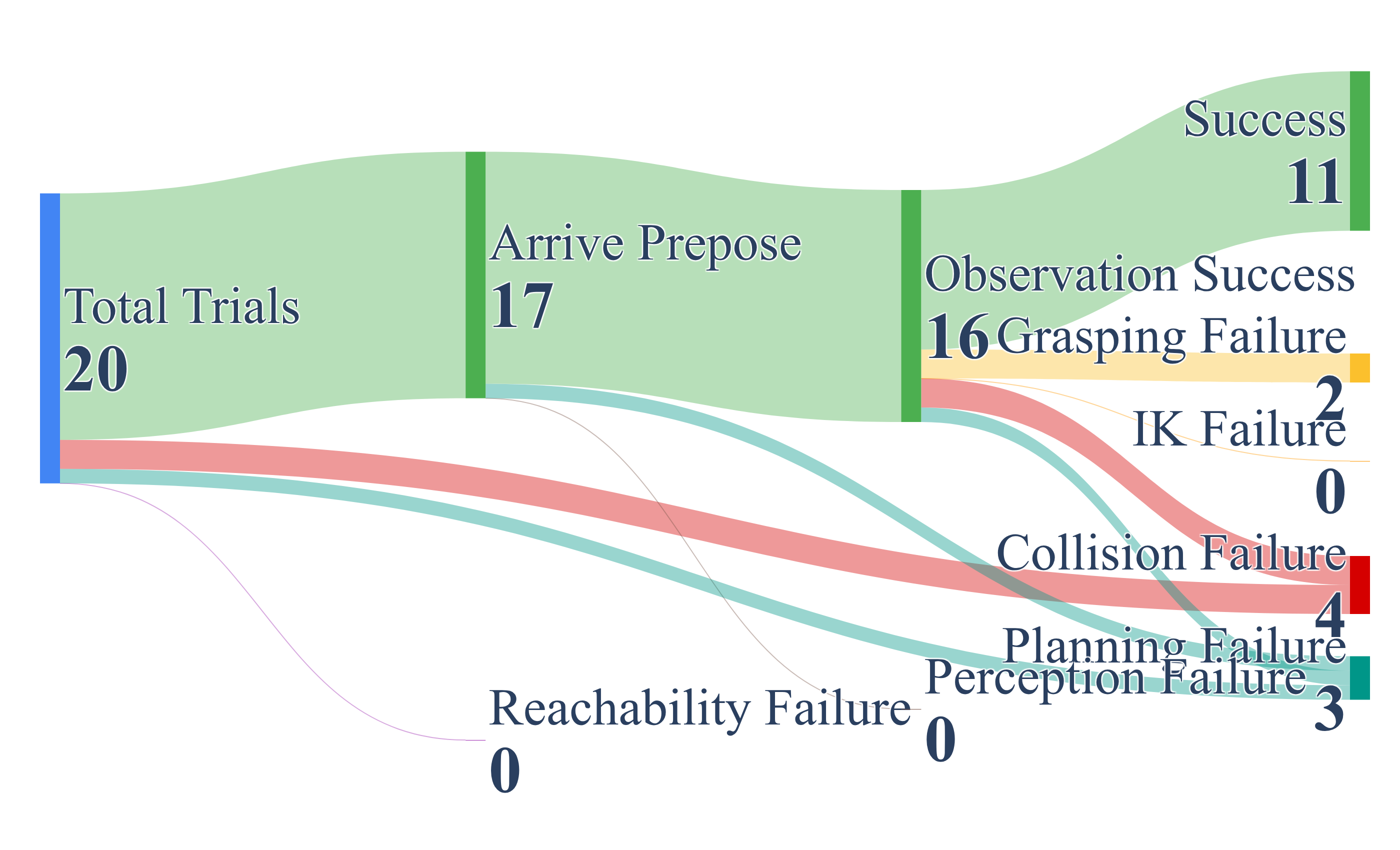}
    \caption{Failure flow in simulation (left two, unknown static and dynamic) and on the real robot (right two, static and dynamic).}
    \label{fig:failure_analysis}
    \label{fig:real_world_results}
    \vspace{-0.4cm}
\end{figure*}

\paragraph{Qualitative Results}
Figure~\ref{fig:execution_sequence} shows a representative rollout.
Notably, the sampled pre-grasp pose avoided self-occlusion and kept the target visible for the final grasp, reflecting the object visibility constraint in action.

\paragraph{Comparison with Baselines}
Our integrated approach outperformed \nam by 16.3 percentage points in unknown and 9.3 percentage points in dynamic environments (Table~\ref{tab:results_sim}).
The decoupled design committed to a base placement before the robot could observe the target's surroundings, which frequently produced unreachable configurations and doubled planning failures, while its fixed-base arm planning struggled in constrained spaces and collided more often.
Removing the subgoal policy entirely (Direct Grasping) caused the largest drop of 43.2 and 41.8 percentage points.
Without the behavior tree, the system could not fall back to repositioning or observation gathering when grasp sampling failed, and planning failures grew by 3.3--5.3$\times$.
Even with the oracle Easy initialization, the learning baselines achieved low success, and SG-VLA remained near zero.
Their failures were dominated by collisions, rising further under the Hard initialization, consistent with acting on visual input without an explicit model of collision geometry.
Neither policy showed signs of seeking out the regions that mattered for safe motion, such as the space it was about to enter.
Such behavior hardly emerges from the learning signal alone, so visibility needs to be treated as an explicit objective.

\paragraph{Ablation Analysis}
Velocity weighting contributed through collision safety.
Removing it more than doubled collisions in static environments and raised them by a third in dynamic ones while other failure categories changed little, so velocity-aware perception improves success almost entirely by avoiding collisions.
The observation stage mattered less, lowering success by 2.1 and 1.6 percentage points, a difference not statistically resolvable over five runs (Welch's $t$-test, $p = 0.20$ static and $p = 0.28$ dynamic).
Path efficiency separates the two designs more clearly, with the full system reaching 62.2\% and 48.0\% SPL against 54.7\% and 45.8\% for the ablation, so without the observation stage the robot succeeds through longer and less direct paths.
Per-scene success rates ranged from 58--82\% in static and 41--67\% in dynamic environments with cross-scene standard deviations of 6.9\% and 7.5\%, reflecting layout and clutter diversity.

\paragraph{Real-World Validation}
We deployed only the full system on the real robot, as baselines lacking active perception and reactive replanning risk damaging the robot and the environment.
It achieved success rates comparable to simulation in both static (65.0\% vs. 70.4\%) and dynamic (55.0\% vs. 57.9\%) environments (Figure~\ref{fig:real_world_results}).
Planning failures primarily came from noisy observations (e.g., floating points in the point cloud) caused by data synchronization issues between the localization system and depth sensors.
Collision failures arose from ground plane filtering that occasionally removed points from low-lying objects like chair legs, and from higher execution noise where the robot got stuck due to friction, then suddenly accelerated, disrupting pose estimation.
Perception failures occurred when segmentation or detection errors led to incorrect grasping.
These factors accounted for the performance gap between simulation and real-world.

\paragraph{Runtime}
Each control cycle took approximately 100\,ms (map update 69\,ms, gaze optimization 26\,ms, collision checking 3\,ms), extending to roughly 150\,ms when replanning was triggered (motion planning 50--80\,ms).
Subgoal sampling (369\,ms) and grasp detection (1.2\,s) ran only at phase transitions, 3.6 and 1.3 times per task.
Velocity-aware gaze added 26\,ms per cycle, the only computational difference from Ours (w/o Velocity Weighting).
The pre-grasp sampler was empirically near complete, with success plateauing near 95\% by 2{,}000 samples, motivating our 2{,}560-sample setting.

\paragraph{Failure Analysis}
Figure~\ref{fig:failure_analysis} and Figure~\ref{fig:success_distribution} show the failure flow and its spatial distribution on our benchmark.
Failures concentrated on objects deep on tables, near walls, and in clutter, where feasible approach directions are scarce.
The primary bottleneck was grasp detection.
Among 46 manually labeled grasp failures, 43.5\% were geometrically sound but kinematically infeasible grasps, 30.4\% were misaligned poses, and 23.9\% slipped during lift.
The dominance of infeasible grasps reveals a domain shift from tabletop to mobile grasping, where the detector ranks candidates by geometric quality alone while reachability depends on base placement and environment constraints, making kinematically informed grasp scoring a clear direction.

Beyond this per-module attribution, the failure cases point to three structural limitations in the coupling between perception and planning.
First, observed free space has limited temporal validity in dynamic scenes.
As the ablation shows, velocity-aware gaze is effective once an obstacle has entered the map.
Inspection of the residual collisions showed that space verified as free was later invalidated by obstacles arriving from adjacent unobserved regions before the gaze revisited it.
Active perception should therefore anticipate where invalidation can originate, for example at occlusion boundaries near the planned route.

Second, reactive map maintenance retains stale evidence.
The map cleared freed space only along observed rays, so removing points required re-observing the same region.
Every dynamic obstacle naturally left such residue. Once the robot observed an obstacle and shifted its gaze, the obstacle departed while its points persisted in the map as floating obstacles, and regions the gaze had no reason to revisit kept them indefinitely, blocking valid paths and surfacing as planning failures.
Removing this failure class requires a predictive environment model that propagates dynamic evidence forward and decays unconfirmed points while preserving occluded static obstacles.

Third, the coupling between perception and planning remains unidirectional.
The gaze policy conditioned on the planned trajectory, while the motion policy optimized collision cost in $M_t$ without considering whether its swept volume could be observed.
Observability is a property of the trajectory itself, determined by head reachability, self-occlusion, and motion direction, with backward motion as the extreme case where the forward-facing camera cannot cover the space being entered.
Visibility should enter motion planning as a cost or constraint, extending joint gaze-motion optimization~\cite{meng2025lookleapplanningsimultaneous} from fixed goals to online refinement under an evolving map.

%% file: conclusion/conclusion.tex
\section{Conclusion}
\label{sec:conclusion}

This paper presented a unified framework for mobile grasping in dynamic, unknown environments under visibility constraints.
By integrating velocity-aware active perception, hierarchical subgoal generation, and fast whole-body replanning, our system improves success rates by 16.3 and 9.3 percentage points over the \nam method and achieves an overall 70.4\% success rate in unknown static environments and 57.9\% in dynamic environments.
In particular, our method reduces collision rates effectively compared to baselines, benefiting from our closed-loop planning and active perception.
These results demonstrate the importance of well-coordinated perception and planning for executing mobile grasping tasks in large-scale dynamic environments.
Real-world experiments demonstrated strong generalization across diverse indoor settings.
Future work follows the lessons identified in our analysis, including predictive environment models, active perception that anticipates unknown dynamics, visibility-aware motion planning, and kinematically informed grasp detection. The modular architecture naturally extends to multi-robot coordination, long-horizon or contact-rich manipulation tasks \cite{huang2026kinder}.

%% file: main.bbl
\begin{thebibliography}{10}
\providecommand{\url}[1]{#1}
\csname url@samestyle\endcsname
\providecommand{\newblock}{\relax}
\providecommand{\bibinfo}[2]{#2}
\providecommand{\BIBentrySTDinterwordspacing}{\spaceskip=0pt\relax}
\providecommand{\BIBentryALTinterwordstretchfactor}{4}
\providecommand{\BIBentryALTinterwordspacing}{\spaceskip=\fontdimen2\font plus
\BIBentryALTinterwordstretchfactor\fontdimen3\font minus
  \fontdimen4\font\relax}
\providecommand{\BIBforeignlanguage}[2]{{%
\expandafter\ifx\csname l@#1\endcsname\relax
\typeout{** WARNING: IEEEtran.bst: No hyphenation pattern has been}%
\typeout{** loaded for the language `#1'. Using the pattern for}%
\typeout{** the default language instead.}%
\else
\language=\csname l@#1\endcsname
\fi
#2}}
\providecommand{\BIBdecl}{\relax}
\BIBdecl

\bibitem{liu2024okrobot}
P.~Liu \emph{et~al.}, ``Ok-robot: What really matters in integrating
  open-knowledge models for robotics,'' \emph{arXiv preprint arXiv:2401.12202},
  2024.

\bibitem{yenamandra2024homerobotopenvocabularymobilemanipulation}
S.~Yenamandra \emph{et~al.}, ``Homerobot: Open-vocabulary mobile
  manipulation,'' 2024.

\bibitem{xiao2025robibutlermultimodalremote}
A.~Xiao \emph{et~al.}, ``Robi butler: Multimodal remote interaction with a
  household robot assistant,'' 2025.

\bibitem{9830863}
F.~Reister, M.~Grotz, and T.~Asfour, ``Combining navigation and manipulation
  costs for time-efficient robot placement in mobile manipulation tasks,''
  \emph{IEEE Robotics and Automation Letters}, vol.~7, no.~4, pp. 9913--9920,
  2022.

\bibitem{6630839}
N.~Vahrenkamp, T.~Asfour, and R.~Dillmann, ``Robot placement based on
  reachability inversion,'' in \emph{2013 IEEE International Conference on
  Robotics and Automation}, 2013, pp. 1970--1975.

\bibitem{somani2013despot}
A.~Somani \emph{et~al.}, ``Despot: Online pomdp planning with regularization,''
  \emph{Advances in neural information processing systems}, vol.~26, 2013.

\bibitem{10611190}
W.~Thomason, Z.~Kingston, and L.~E. Kavraki, ``Motions in microseconds via
  vectorized sampling-based planning,'' in \emph{2024 IEEE International
  Conference on Robotics and Automation (ICRA)}, 2024, pp. 8749--8756.

\bibitem{Bajracharya_2023}
M.~Bajracharya \emph{et~al.}, ``Demonstrating mobile manipulation in the wild:
  A metrics-driven approach,'' in \emph{Robotics: Science and Systems XIX},
  ser. RSS2023.\hskip 1em plus 0.5em minus 0.4em\relax Robotics: Science and
  Systems Foundation, 2023.

\bibitem{9517662}
Y.~Meng, Y.~Chen, and Y.~Lou, ``Uncertainty aware mobile manipulator platform
  pose planning based on capability map,'' in \emph{2021 IEEE International
  Conference on Real-time Computing and Robotics (RCAR)}, 2021, pp. 123--128.

\bibitem{xiao2026hume}
A.~Xiao \emph{et~al.}, ``Hypothesis-driven model expansion under uncertainty
  for open-world robot planning,'' in \emph{Robotics: Science and Systems
  (RSS)}, 2026.

\bibitem{yu2026canine}
C.~Yu \emph{et~al.}, ``Canine: Coaching visually impaired users for interactive
  navigation with a robot guide dog,'' in \emph{Robotics: Science and Systems
  (RSS)}, 2026.

\bibitem{Covic_2025}
N.~Covic \emph{et~al.}, ``Real-time sampling-based safe motion planning for
  robotic manipulators in dynamic environments,'' \emph{IEEE Transactions on
  Robotics}, vol.~41, p. 5287–5306, 2025.

\bibitem{Finean_2021}
M.~N. Finean, W.~Merkt, and I.~Havoutis, ``Simultaneous scene reconstruction
  and whole-body motion planning for safe operation in dynamic environments,''
  in \emph{2021 IEEE/RSJ International Conference on Intelligent Robots and
  Systems (IROS)}.\hskip 1em plus 0.5em minus 0.4em\relax IEEE, 2021, pp.
  3710--3717.

\bibitem{goretkin2018look}
G.~Goretkin, L.~P. Kaelbling, and T.~Lozano-P{\'e}rez, ``Look before you sweep:
  Visibility-aware motion planning,'' in \emph{Proceedings of the Workshop on
  the Algorithmic Foundations of Robotics (WAFR)}.\hskip 1em plus 0.5em minus
  0.4em\relax Springer, 2018, pp. 1--16.

\bibitem{ichter2017perceptionawaremotionplanningmultiobjective}
B.~Ichter \emph{et~al.}, ``Perception-aware motion planning via multiobjective
  search on gpus,'' 2017.

\bibitem{meng2025lookleapplanningsimultaneous}
Q.~Meng \emph{et~al.}, ``Look as you leap: Planning simultaneous motion and
  perception for high-dof robots,'' 2025.

\bibitem{4209173}
K.~E. Bekris and L.~E. Kavraki, ``Greedy but safe replanning under kinodynamic
  constraints,'' in \emph{Proceedings 2007 IEEE International Conference on
  Robotics and Automation}, 2007, pp. 704--710.

\bibitem{breyer2022closedloopnextbestviewplanningtargetdriven}
M.~Breyer \emph{et~al.}, ``Closed-loop next-best-view planning for
  target-driven grasping,'' 2022.

\bibitem{10.1109/ICRA.2019.8793805}
D.~Morrison, P.~Corke, and J.~Leitner, ``Multi-view picking: Next-best-view
  reaching for improved grasping in clutter,'' in \emph{2019 International
  Conference on Robotics and Automation (ICRA)}.\hskip 1em plus 0.5em minus
  0.4em\relax IEEE Press, 2019, p. 8762–8768.

\bibitem{jauhri2024activeperceptivemotiongenerationmobile}
S.~Jauhri, S.~Lueth, and G.~Chalvatzaki, ``Active-perceptive motion generation
  for mobile manipulation,'' 2024.

\bibitem{marques2025mapspacebeliefprediction}
J.~M.~C. Marques \emph{et~al.}, ``Map space belief prediction for
  manipulation-enhanced mapping,'' 2025.

\bibitem{Haviland_2021}
J.~Haviland and P.~Corke, ``Neo: A novel expeditious optimisation algorithm for
  reactive motion control of manipulators,'' \emph{IEEE Robotics and Automation
  Letters}, vol.~6, no.~2, p. 1043–1050, 2021.

\bibitem{Mukadam_2018}
M.~Mukadam \emph{et~al.}, ``Continuous-time gaussian process motion planning
  via probabilistic inference,'' \emph{The International Journal of Robotics
  Research}, vol.~37, no.~11, pp. 1319--1340, 2018.

\bibitem{10610192}
C.~Wu \emph{et~al.}, ``Real-time whole-body motion planning for mobile
  manipulators using environment-adaptive search and spatial-temporal
  optimization,'' in \emph{2024 IEEE International Conference on Robotics and
  Automation (ICRA)}, 2024, pp. 1369--1375.

\bibitem{ahn2022do}
M.~Ahn \emph{et~al.}, ``Do as i can, not as i say: Grounding language in
  robotic affordances,'' in \emph{Conference on Robot Learning}.\hskip 1em plus
  0.5em minus 0.4em\relax PMLR, 2022, pp. 287--315.

\bibitem{huang2023voxposer}
W.~Huang \emph{et~al.}, ``Voxposer: Composable 3d value maps for robotic
  manipulation with language models,'' in \emph{Conference on Robot
  Learning}.\hskip 1em plus 0.5em minus 0.4em\relax PMLR, 2023, pp. 540--562.

\bibitem{jauhri2022robot}
S.~Jauhri, J.~Peters, and G.~Chalvatzaki, ``Robot learning of mobile
  manipulation with reachability behavior priors,'' \emph{IEEE Robotics and
  Automation Letters}, vol.~7, no.~3, p. 8399–8406, 2022.

\bibitem{zhang2024gammagraspabilityawaremobilemanipulation}
J.~Zhang \emph{et~al.}, ``Gamma: Graspability-aware mobile manipulation policy
  learning based on online grasping pose fusion,'' 2024.

\bibitem{intelligence2025pi05visionlanguageactionmodelopenworld}
P.~Intelligence \emph{et~al.}, ``$\pi_{0.5}$: a vision-language-action model
  with open-world generalization,'' 2025.

\bibitem{tu2026sgvlalearningspatiallygroundedvisionlanguageaction}
R.~Tu \emph{et~al.}, ``Sg-vla: Learning spatially-grounded
  vision-language-action models for mobile manipulation,'' 2026.

\bibitem{wu2025momanipvlatransferringvisionlanguageactionmodels}
Z.~Wu \emph{et~al.}, ``Momanipvla: Transferring vision-language-action models
  for general mobile manipulation,'' 2025.

\bibitem{zitkovich2023rt}
B.~Zitkovich \emph{et~al.}, ``Rt-2: Vision-language-action models transferred
  from web-scale real-world data,'' \emph{arXiv preprint arXiv:2307.15818},
  2023.

\bibitem{finean2021where}
M.~Finean, W.~Merkt, and I.~Havoutis, ``Where should i look optimised gaze
  control for whole-body collision avoidance in dynamic environments,''
  \emph{IEEE Robotics and Automation Letters}, vol.~PP, pp. 1--1, 2021.

\bibitem{sundermeyer2021contact-graspnet}
M.~Sundermeyer \emph{et~al.}, ``Contact-graspnet: Efficient 6-dof grasp
  generation in cluttered scenes,'' in \emph{2021 IEEE International Conference
  on Robotics and Automation (ICRA)}.\hskip 1em plus 0.5em minus 0.4em\relax
  IEEE Press, 2021, p. 13438–13444.

\bibitem{beeson2015trac-ik}
P.~Beeson and B.~Ames, ``Trac-ik: An open-source library for improved solving
  of generic inverse kinematics,'' in \emph{2015 IEEE-RAS 15th International
  Conference on Humanoid Robots (Humanoids)}.\hskip 1em plus 0.5em minus
  0.4em\relax IEEE Press, 2015, p. 928–935.

\bibitem{Kurzer1057261}
K.~Kurzer, ``Path planning in unstructured environments : A real-time hybrid a*
  implementation for fast and deterministic path generation for the kth
  research concept vehicle,'' Master's thesis, KTH, Integrated Transport
  Research Lab, ITRL, 2016.

\bibitem{844730}
J.~Kuffner and S.~LaValle, ``Rrt-connect: An efficient approach to single-query
  path planning,'' in \emph{Proceedings 2000 ICRA. Millennium Conference. IEEE
  International Conference on Robotics and Automation. Symposia Proceedings
  (Cat. No.00CH37065)}, vol.~2, 2000, pp. 995--1001 vol.2.

\bibitem{Geraerts2007}
R.~Geraerts and M.~H. Overmars, ``Creating high-quality roadmaps for motion
  planning in virtual environments,'' in \emph{2006 IEEE/RSJ International
  Conference on Intelligent Robots and Systems}, 2006, pp. 4355--4361.

\bibitem{Hauser2010}
K.~Hauser and V.~Ng-Thow-Hing, ``Fast smoothing of manipulator trajectories
  using optimal bounded-acceleration shortcuts,'' in \emph{2010 IEEE
  International Conference on Robotics and Automation}, 2010, pp. 2493--2498.

\bibitem{Pan2012}
J.~Pan, L.~Zhang, and D.~Manocha, ``Collision-free and smooth trajectory
  computation in cluttered environments,'' \emph{The International Journal of
  Robotics Research}, vol.~31, no.~10, pp. 1155--1175, 2012.

\bibitem{taomaniskill3}
S.~Tao \emph{et~al.}, ``Maniskill3: Gpu parallelized robotics simulation and
  rendering for generalizable embodied ai,'' \emph{Robotics: Science and
  Systems}, 2025.

\bibitem{szot2021habitat}
A.~Szot \emph{et~al.}, ``Habitat 2.0: Training home assistants to rearrange
  their habitat,'' in \emph{Advances in Neural Information Processing Systems
  (NeurIPS)}, 2021.

\bibitem{eppner2021acronym}
C.~Eppner, A.~Mousavian, and D.~Fox, ``Acronym: A large-scale grasp dataset
  based on simulation,'' in \emph{2021 IEEE International Conference on
  Robotics and Automation (ICRA)}, 2021, pp. 6222--6227.

\bibitem{huang2026kinder}
Y.~Huang \emph{et~al.}, ``Kinder: A physical reasoning benchmark for robot
  learning and planning,'' in \emph{Robotics: Science and Systems (RSS)}, 2026.

\end{thebibliography}
